\documentclass[a4paper, 10pt, conference]{ieeeconf}      
\IEEEoverridecommandlockouts                              
\usepackage{graphics} 
\usepackage{subcaption}
\usepackage{cite}
\usepackage{epsfig} 
\usepackage{mathptmx} 
\usepackage{times} 
\usepackage{amsmath,amsfonts} 
\usepackage{amssymb}  
\usepackage{bm} 

\usepackage{xcolor}

\usepackage{cite}
\usepackage{gensymb}
\title{\LARGE \bf
Distributed Cascade Force Control of Soft-Tactile-Based Multi-robot System for Object Transportation
}

\author{Duy Anh Nguyen$^{1}$, 
    Nhat Minh Dinh Le$^{2}$,
    Nhan Huu Nguyen$^{2}$, Pham Duy Hung$^{3}$, 
    \\
    Van Anh Ho$^{2, 4}$
    and Trung Dung Ngo$^{1}$
\thanks{Trung Ngo acknowledges the support of the Natural Sciences and Engineering Council of Canada, NSERC-RGPIN-2023-05961, DGDND-2023-05961, and NSERC-CREATE 528099-2019.}
\thanks{$^{1}$D.A. Nguyen and T.D. Ngo are with the More-Than-One Robotics Laboratory, Faculty of Sustainable Design Engineering, University of Prince Edward Island, Charlottetown, Canada.}%
\thanks{$^{2}$N.M.D. Le, N.H. Nguyen, and V.A. Ho are with the Soft Haptics Lab., School of Materials Science, Japan Advanced Institute of Science and Technology, Asahidai 1-1, Nomi, Ishikawa, 923-1292 Japan.}%
\thanks{$^{3}$P.D. Hung is with University of Engineering and Technology, Vietnam.}
\thanks{$^{4}$V.A. Ho is with Japan Science and Technology Agency, PRESTO, Kawaguchi, Saitama, 332-0012 Japan.}%
\thanks{$^{*}$Corresponding author. Email: {\tt\small  tngo@upei.ca}}%
}

\begin{document}

\maketitle
\thispagestyle{empty}
\pagestyle{empty}

\begin{abstract}
In this paper, we present a \underline{d}istributed \underline{c}ascade \underline{f}orce \underline{c}ontrol system (DCFC) for multiple robots with the aim of pushing a rigid object towards a desired moving target without their inter-robot communication. These mobile robots are equipped with 360-degree vision-based soft tactile sensors utilized to determine contact location and resultant impact force. 
By investigating the dynamics of moving rigid objects on the flat, we proposed a distributed cascade control. The inner loop control incorporates contact force and positioning, ensuring the robots' pushing contact and application of the desired force to the object. The outer loop control coordinates the robots to push the object in a desired direction without inter-robot communication, regardless of the unknown object mass. The stability and convergence of the system are verified using the Lyapunov stability theory. We also conducted simulation and real-word experiments to validate the performance of the proposed control method, and the experimental results showcase the successful coordination of multiple robots in pushing an object towards a moving desired direction.

\end{abstract}

\section{INTRODUCTION}
\begin{figure}[t!]
\centerline{\includegraphics[width=\linewidth]{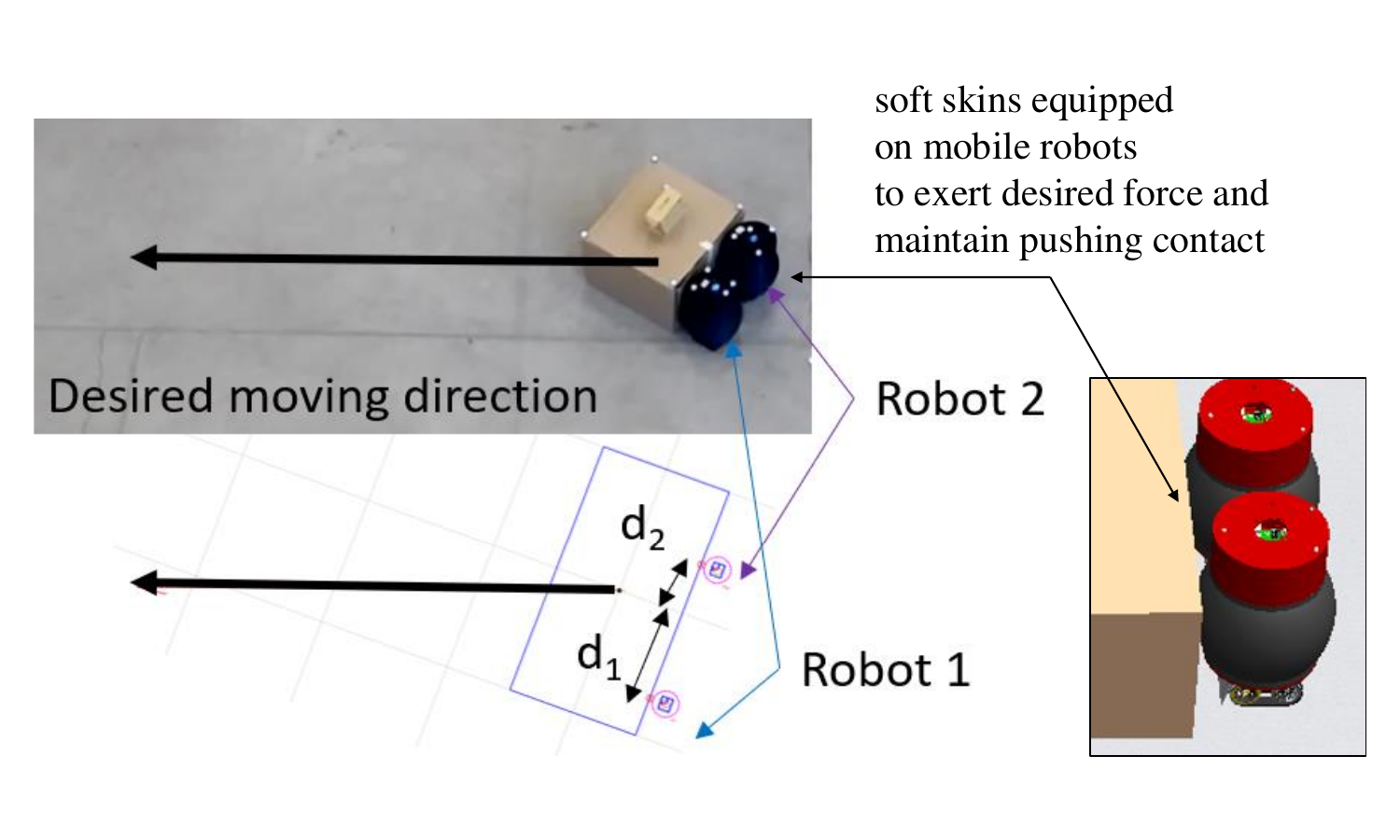}}
\caption{Real-world experiment (top) and simulation (bottom) of multiple soft-tactile-based mobile robots performing an object transportation task.}
\label{fig_simu}
\end{figure}

Collective object transportation by multiple robots can be employed in many practical applications, such as smart warehouses, construction sites, and search and rescue missions. To achieve collaborative tasks, a distributed control system for consensus decision-making among the robots conducted at the individual level based on local information and interactions with neighbouring robots is required. The advantages of a distributed control system for multi-robot systems are known as scalability, robustness, and adaptability when deployed in dynamic and uncertain environments. This allows the robotic system to handle the complexity of a given task and the number of robots required for the task, such as heavy object transportation, inspired by the ants as illustrated in Fig. \ref{fig_simu}. 


Research on object transportation by multi-robot systems is roughly categorized into two groups named \emph{pushing} and \emph{grasping}. In general, the pushing method involves exerting force on an object through physical contact but physical attachment, while the grasping approach requires a physical connection between the robot and the transported object. In general, pushing is more challenging than grasping because no additional gripping tool is needed while the dynamics of rigid objects are critically required.

The primitive strategy for multiple mobile robots to transport an object involves controlling robots' velocity and/or formation while dealing with forces between the object, robots, and the surface as disturbances. Inspired by collective behaviours in social insects in \cite{kube1}, the authors proposed simple group behaviour-based algorithms for pushing a box forward without communication. These algorithms were extended by employing sensors to detect the goal, allowing the robots to relocate around and push the object when the goal was detected \cite{kube2}. Further studies analyzed the stagnation recovery based on ants' behavior and tested the box-pushing algorithm by tracking changing goals with different types of boxes \cite{kube3,kube4}. However, these kinetic strategies have limitations as they did not account for the physical interaction between the object and the surface, which plays a critical role in object transportation, especially across various terrains or slopes \cite{caging}. Uncontrollable forces between the object and the robots can exceed safety limits and harm both the robots and the object, particularly fragile ones. These methods may lead to inaccuracy in task performance as they do not incorporate rigid-body object dynamics to represent object behaviour.

Recent studies addressed the above-mentioned problems by considering \emph{object dynamics} and \emph{robot forces}. Grasping techniques with controllable exerted force provided greater control freedom for robots to push or pull objects with the motion of local objects measured by each robot as implicit information for force coordination \cite{Fants,Stan_1_1,Stan_1_2}. In \cite{Fants}, it was proved that all the robots' contact forces converged to a leader force direction, resulting in a force-amplifying effect. An extension of this work considered both force and torque consensus \cite{Stan_1_1}, as well as steering the leader's force to navigate the system in narrow spaces \cite{Stan_1_2}. In other directions, decentralized adaptive control leveraged the theory of adaptive control method to estimate system parameters and manipulate the object simultaneously \cite{DAC,DAC2}. The desired object states, such as linear and angular velocity and desired trajectories, were achieved and proved through the Lyapunov stability theory. Within the scope of pushing methods, some researchers proposed collective transportation strategies based on occlusion, where robots only pushed the object when their vision of the goal was obscured \cite{Shef}, \cite{Shef2}. Maintaining continuous pushing on the object while transporting it towards the desired direction is always challenging with pushing methods. In \cite{neumann}, object transportation was performed by employing explicit force control of two pushing robots while cluster space control for the formation and relative positioning was maintained through continuous intercommunication. 


This paper introduces a distributed cascade control system for multiple soft-tactile-based robots specialized in transporting rectangle-shaped objects without communication. Using the soft tactile sensors developed in our previous work \cite{ViART}, real-time tactile information on each robot is obtained from physical interaction within a multi-agent robotic system and with the surroundings. The proposed control strategy enables the robots to exert pushing forces while maintaining constant contact with the object. The proposed approach demonstrates adaptability and versatility in cascade distributed control for soft-tactile-based robot systems, accommodating various object masses and dealing with moving in the desired direction. 

The paper is organized as follows. Section II overviews rigid-body object dynamics and introduces our mobile robot platform. Section III outlines the problem and objectives. Section IV examines our proposed DCFC and includes proof of its stability. Section V presents the results and the discussion of both simulation and real-world experiments. Finally, the conclusions are presented in Section VI.  

 \section{PRELIMINARIES}
\subsection{Modeling Rigid-Body Object Dynamics and Pushing Forces}

In this section, we consider a rigid-body object capable of both translation and rotation in a 2D frictional environment as described in Fig.~\ref{fig_force}. The translational velocity and rotational velocity of the object are represented by $\mathbf{{v}}_G \in \mathbb{R}^2$, and $\boldsymbol{\omega}(t)$, respectively. To describe these motions under the influence of $n$ external pushing forces $\bm{f}_i \in \mathbb{R}^2$, we use Newton's law of motion and Euler's rigid-body equation, respectively, as:
\begin{equation}
	m \mathbf{\dot{v}}_G = \sum_{i=1}^n \bm{f}_i + \bm{F}_{f}, \label{eq_motion3}
\end{equation}
\begin{equation}
	I\bm{\dot{\omega}} =  \sum_{i=1}^n{(\bm{r}_i \times \bm{f}_i)} + \bm{M}_{f}, \label{eq_motion4}
\end{equation}
where $m$ and $I$ are denoted as the object's mass and inertia mass, respectively. $\bm{F}_{f}$ and $\bm{M}_{f}$ representing the frictional force and torque between the object's surface and the ground are modelled as zeroth and first-order functions of motion, respectively:

\begin{figure}[t!]h
		\centerline{\includegraphics[width=0.75\linewidth]{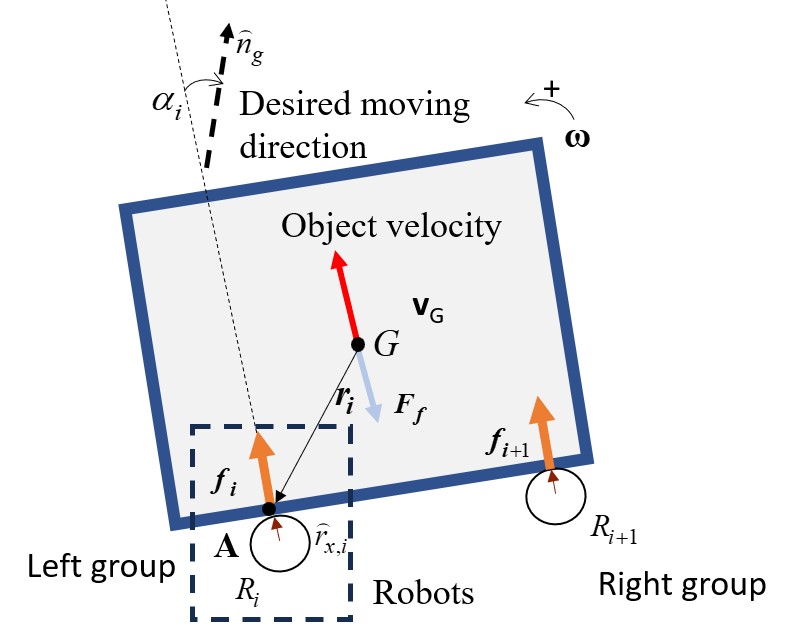}}
		\caption{Free-body diagram of an object under a pushing force $\bm{f}_i$ .}
		\label{fig_force}
	\end{figure}

\begin{equation}
	\bm{F}_f = -\mu_{0,t}m \bm{\sigma(\dot{v}_G)}  -\mu_{1,t} \bm{v}_G \label{eq_fric1},
\end{equation}
\begin{equation}
	\bm{M}_f = -\mu_{0,r}I \bm{\sigma(\omega)} - \mu_{1,r}\bm{\omega} \label{eq_fric2},
\end{equation}
where $\bm{\sigma(\dot{x})} = \bm{\dot{x} / |\dot{x}|}$ represents the sign function, and $\mu_{0,t}$, $\mu_{1,t}$, $\mu_{0,r}$, and $\mu_{1,r}$ are the friction parameters accordingly.

\subsection{Visual-based Tactile Sensor Modeling}
Our mobile robot platform, developed as in \cite{ViART}, is equipped with a 360-degree polymer material capable of deformation and exerting pushing force on objects. When force is applied to the tactile sensor, 24 markers located inside are displaced and captured by the fisheye camera mounted on the center of the sensor. Through a pre-calibration system, each robot can accurately estimate the applied force $f_i$ and contact angle $\theta_i$ with high precision (the mean absolute error of 0.12 $N$ and 1.12$^\circ$, respectively). We approximate the robot's shape as a circle with a radius to model the pushing force exerted by the robot.
Within the scope of this paper, we simplify the force magnitude model as a simple spring model that is applied for control system design:
  \begin{equation}
 \boldsymbol{f}_i(\Delta_i) = k\boldsymbol{\Delta}_i,
 \label{eq_spring}
\end{equation}
where $\boldsymbol{\Delta}_i \in \mathbb{R}^2$ is a deformation distance ($m$), and $k$ is a constant, which is estimated at around 400 ($N/m$) in this study. Note that, by the third Newton law, the strength of applied perpendicular force to the robot surface equals the perpendicular pushing force to the object.    
Denote $A_i$ as the nearest point of robot $i$ to the object, we have: 
\begin{equation}
 \boldsymbol{\Delta}_i= \boldsymbol{x}_{Ai} - \boldsymbol{x}_i. \label{eq_deform}
\end{equation}
\begin{figure}[t!]
		\centerline{\includegraphics[width=0.5\linewidth]{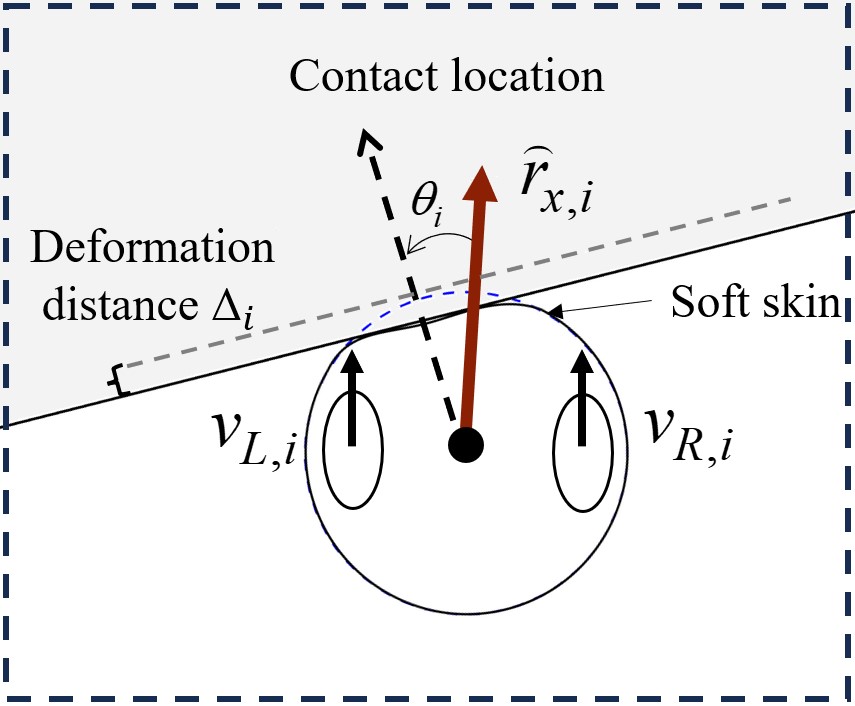}}
		\caption{The Visual-based tactile model with contact angle and contact force. }
		\label{fig_background2}
	\end{figure}
 \subsection{Interaction force between robot and object surface}	
When a robot touches or pushes an object, there exists an interactive force between the robot and the object's surface. This force includes both the perpendicular force to the object's surface and the frictional force between them. It is important to note that this friction prevents the sliding between the tactile sensor and the object's surface. Furthermore, if there is minimal relative motion between the robot and the object (i.e., $\mathbf{v}_i \approx \mathbf{v}_G$), especially in the steady state when $\omega \approx 0$, the friction between the two surfaces can be ignored.
\section{PROBLEM DESCRIPTION}
Consider a set of $n$ soft-tactile-based mobile robots with a given task of transporting a rectangular-shaped rigid object of indeterminate mass and inertia toward a specified direction, denoted as $\widehat{n_g}$. Each robot is capable of measuring the relative orientation $\alpha_i \in [-\pi, \pi]$ between its heading $\widehat{r}_{x,i}$ and $\widehat{n_g}$. Furthermore, each robot can estimate the pushing force $f_i$ exerted on the object. Given their limited sensory capabilities, it is assumed that these pushing forces act perpendicularly to the object's surface. In the initiation of the pushing phase, two distinct robot groups are designated. Both groups are situated on the same edge of the object so that their relative angles satisfy $\alpha_i \in [-\frac{\pi}{2}, \frac{\pi}{2}]$. These groups are positioned on opposing sides of the projection of the center of mass, $G$, on the above edge: $n_L$ robots are on the left, and $n_R$ robots are on the right. The positioning of these groups can result from user intervention or emerge from simple behavior-based control of robots, which may or may not require inter-robot communication.

The primary challenge is to enable the robots to consistently maintain pushing contact with the object, thus providing stability to the entire system of the object and robots. This problem can be divided into two key objectives:
\begin{itemize}
\item (1) To synchronize the object's velocity along with the targeted direction.
\item (2) To make each robot maintain its pushing contact and apply the necessary force to facilitate the object's transportation.
\end{itemize}

The direction orthogonal to $\widehat{n_g}$ is represented by $\widehat{n_p}$. In scenarios where the object's velocity is not measurable, the first objective can be redefined: the pushing force vectors should be in alignment with the targeted direction, and the total robot force in $\widehat{n_p}$ should be zero:

\begin{align}
 \sum_{i=1}^n \bm{f}_i \cdot \widehat{n_g} &=  \sum_{i=1}^n f_i  > 0, \label{eq_ng} \\
 \sum_{i=1}^n \bm{f}_i \cdot \widehat{n_p} &= 0. \label{eq_np}
\end{align}
Under these conditions, the object's velocity can be adjusted to the desired direction.

By projecting Newton's second law given by Eq. \ref{eq_motion3} onto the $\widehat{n_p}$ axis, we derive:
\begin{equation}
m \mathbf{\dot {v}}_G \cdot \widehat{n_p}  = \sum_{i=1}^n \bm{f}_i \cdot \widehat{n_p} + \bm{F}_{f} \cdot \widehat{n_p}. \label{eq_motion9}
\end{equation}
Incorporating Eqs. \ref{eq_fric1}, \ref{eq_np}, and \ref{eq_motion9}, we establish:
\begin{equation}  \bm{\dot {v}}_G \cdot \widehat{n_p}  + \left(\frac {\mu_{1,t}}{m} +\frac{\mu_{0,t}}{\left\|v_G\right\|}\right) \bm{v}_G \cdot \widehat{n_p} = 0. 
\end{equation}
Given a frictional environment, where $\mu_{0,t}+\mu_{1,t} >0$, it follows that $\bm{\dot {v}}_G \cdot \widehat{n_p}$ converges toward zero. Similarly, the object's velocity toward $\widehat{n_g}$ stabilizes at a constant magnitude if the $\sum_{i=1}^n f_i$ is constant.

\section{DISTRIBUTED CASCADE FORCE CONTROL FOR SOFT-TACTILE-BASED ROBOTS (DCFC)}
With two given objectives, we propose a DCFC control consisting of two control systems equivalent to the following sub-objectives:
\begin{figure*}[t!]
		\centerline{\includegraphics[width=0.64\linewidth]{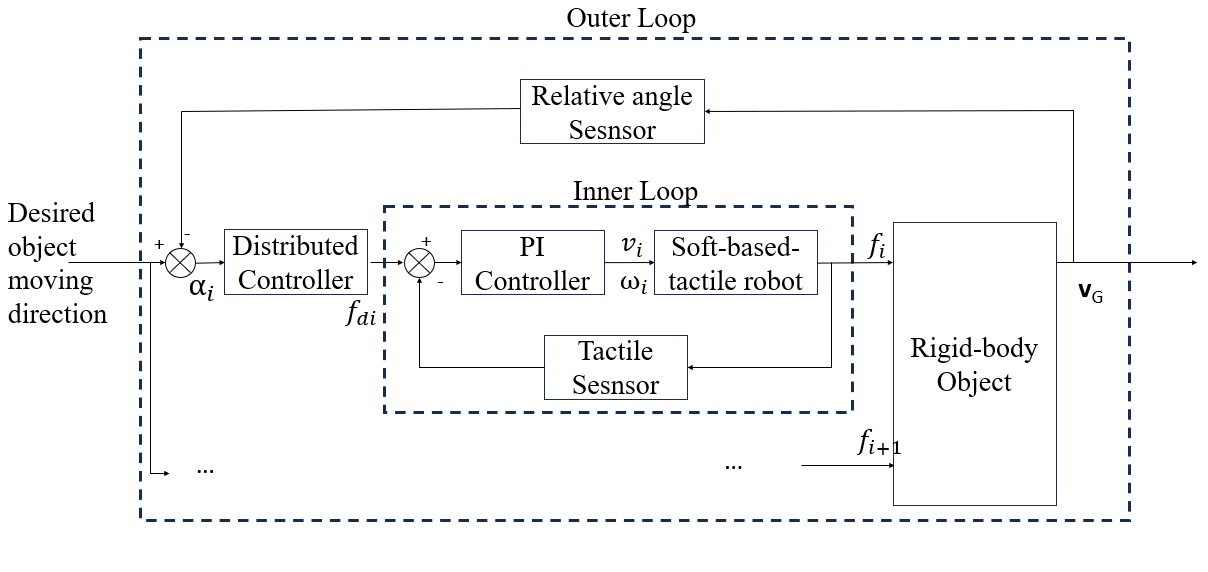}}
		\caption{ Distributed Cascade control (DCFC) graph including an inner loop to maintain pushing contact and force, and an outer loop to align force vector to the desired direction.}
		\label{fig_cascade}
	\end{figure*}
 
\begin{itemize}
\item The outer loop is distributed control with the force policy that is applied to stabilize object rotation and make the pushing force direction aligned with the desired direction. 
\begin {equation}
\lim_{t\to\infty} \omega (t)  = 0. \label{eq_out1}
\end{equation}

 \begin {equation}
 \lim_{t\to\infty} \alpha_i(t) = 0 , \forall i \in n. \label{eq_out2}
\end{equation}
\item The inner loop is a multiple input multiple output (MIMO) control that is used to maintain pushing contact and exert the desired force $f_{di}$ from the above policy.
 \begin {equation}
 \lim_{t\to\infty} f_i  = f_{di}, \forall i \in n. 
 \label{eq_in1}
\end{equation}
 \begin {equation}
 \lim_{t\to\infty} \theta_i  = 0, \forall i \in n. 
 \label{eq_in2}
 \end{equation}
\end{itemize}
The cascade control graph is described in Fig. \ref{fig_cascade}.
\subsection{Outer loop control - Distributed force policy making force direction aligned with the desired direction}
We proposed a distributed control policy for each robot in two groups, named left group and right group, respectively, as follows:
\begin{equation}
		f_{j}^{L}(t)= \frac{1}{n_L}\left( f_0 -  \left(K_{Po}\alpha_{i}(t) + K_{Io} \int_0^t\alpha_{i}(t) +K_{Do}\dot{\alpha}_{i}(t)\right) \right)\label{eq_f1}, 
	\end{equation}
	\begin{equation}
		f_{k}^{R}(t)= \frac{1}{n_R}\left( f_0 +K_{Po}\alpha_{i}(t) + K_{Io} \int_0^t\alpha_{i}(t) +K_{Do}\dot{\alpha}_{i}(t) \right) \label{eq_f2}.
	\end{equation}
where $f_0$, $K_{Po}$, and $K_{Io}$ are the control parameters of the outer loop.

\textit{Stability proof}: For a rectangular-shaped object, robots on the left group always produce a negative moment, and robots on the right group produce a positive moment. The moment produced by robot $i$ can be written as:
\begin{equation}
\bm{r}_i \times \bm{f}_i = f_id_i, \label{eq_omega2}
\end{equation}
where $d_i$ is the lever arm of the pushing force, $d_i=d_{j}^{L}<0 $ if the robot is on the left edge, and $d_i=d_{k}^{R}>0 $ if the robot is on the right edge.
 Moreover, the rate of the relative angle measured by each robot is equal to the angular rate of the object since the desired distance is slowly changed or even constant.
 \begin{equation}
\dot{\alpha_i}= \dot{(\widehat{n}_g,\widehat{r}_{x,i})} \approx -\omega,\forall i \in n. \label{eq_omega}
\end{equation}

 From Eqs. \ref{eq_motion4}, \ref{eq_fric2}, \ref{eq_f1}, \ref{eq_f2}, \ref{eq_omega2}, \ref{eq_omega}, we derive: 
 \begin{equation}
		{\ddot{\alpha_i}} + \left({\frac{\mu_{0,r} }{\left\|\dot{\alpha_i}\right\|}} + \frac{\mu_{1,r}}{I} + g(d_i)K_{Do} \right)\dot{\alpha_i}+   g(d_i)K_{Po}\alpha_i + b(f_0,d_i)=0  \label{eq_conv}
	\end{equation}
 where $g(d_i)=\frac{1}{n_L}\sum_{j=1}^{n_L}{|d^L_j|} + \frac{1}{n_R}\sum_{k=1}^{n_R}{|d^R_k|} >0$,
 $b(f_0,d_i)= f_0(\frac{1}{n_L}\sum_{j=1}^{n_L}d_j^L+\frac{1}{n_R}\sum_{k=1}^{n_R}d_k^R) + \int_0^t\alpha_{i}(t)g(d_i)K_{Io}$. If the object mass is uniform and the number of robots in each group is equal, $ \frac{1}{n_L}\sum_{j=1}^{n_L}d_j^L +\frac{1}{n_R}\sum_{k=1}^{n_R}d_k^R  \approx 0$. In other cases, the integral part is added to compensate for the constant moment $f_0(\frac{1}{n_L}\sum_{j=1}^{n_L}d_j^L+\frac{1}{n_R}\sum_{k=1}^{n_R}d_k^R)$. From Eq. \ref{eq_conv}, if all of the coefficients are greater than zero and $b(f_0,d_i)\approx 0$, the system is stabilized based on the global invariant set theorem with Lyapunov stability approach \cite{slotine}, the angle error or relative angle $\alpha_i$ is converged to 0 with the rate that can be adjusted by $K_{Po}$ and $K_{Do}$.
 
\subsection{Inner loop control - Distributed contact maintenance and force control}
In the beginning, the skin of the robots is already in contact with the object, and the friction between the object and the robots helps minimize sliding between the robot and the object. However, rolling without slipping can occur when the object is moving. 
In the differential drive robot model, two components, including transitional and rotational behaviour, are required as $v_i$ and $\omega_i$, respectively. The rolling motion from the robot's heading is represented by contact angle $\theta_i$, while the transitional motion helps the robots maintain a certain deformation distance to exert the desired force. We propose a control law for the inner loop:
\begin{equation}
\omega_i= K_{\omega}\theta_i,
\end{equation}
\begin{equation}
e_{fi}(t)= f_{di}(t)-f_i(t),
\label{eq_error}
\end{equation}
\begin{equation}
v_i(t)= K_{P,in}e_{fi}(t) + K_{I,in}\int_0^te_{fi}(t),
\label{eq_controllaw}
\end{equation}
where $K_\omega$, $K_{P,in}$, and $K_{I,in}$ are the control parameters of the inner loop.
Consequently, the differential drive kinematic equations for the left and right wheel velocities are as follows:
\begin{equation}
v_{Li}= v_i - \frac{\omega_i L}{2} \text{  and  } v_{Ro}= v_i + \frac{\omega_i L}{2}.
\end{equation}

\textit{Stability proof:} From Eqs. \ref{eq_spring} and \ref{eq_deform}, the model of the robot force magnitude can be described as:
\begin{equation}
\dot{f}_i=k(v_{Ai}-{v}_i).
\label{eq_forcemodel}
\end{equation}
Taking the derivative of the Eq.~ \ref{eq_forcemodel}, substituting the control law Eq.~\ref{eq_controllaw}, force error Eq. \ref{eq_error}, and assuming that $v_{Ai}$ reaches constant over a period of time, we derive: 
\begin{equation}
\ddot{e}_{fi}+kK_{P,in}\dot{e}_{fi}+ kK_{I,in}e_{fi} =0.
\end{equation}
The force error $e_{fi}$ converges to zero with the rate that can be adjusted by $K_{P,in}$ and $K_{I,in}$.

\section{EXPERIMENT RESULT AND DISCUSSION}

 We present the effectiveness of our proposed DCFC in both simulation and real-world experiments with a group of differential-drive tactile-based mobile robots transporting a rectangular-shaped object, as shown in Fig. \ref{fig_simu}. The object and environment parameters used for simulation are described in Table \ref{table:cp1}. Object mass $m$ and kinetic friction $\mu_{0,1}$ in the real-world experiment are approximated as in the simulation. The parameters we used in the simulation and real-world experiment are described in Table \ref{table:cp2}. The PI control parameters are tunned by Ziegler-Nichol's second method for the inner loop first, then the outer loop. The offset force $f_0$ in this scope is set to be $1.5N$ for each robot to overcome static friction ($ \left\|\sum_{i=1}^{n} \bm{f}_i\right\|=n*f_0 >  mg\mu_{0,t}$ as $ 3> 1.64 (N)$). Note that, it is feasible for each robot to adaptively find $f_0$ at the beginning of the pushing phase by an integral control with the robot's velocity feedback.
\begin{table}[t!]
    \centering
    \caption{Object and environment Parameters}
    \begin{tabular}{| c | c | c | c | c | c | c |} 
        \hline
        \textbf{Parameters} & $m$ & $I$ & $\mu_{0,t}$ & $\mu_{1,t}$ & $\mu_{0,r}$ & $\mu_{1,r}$ \\    
        \hline
        \textbf{Scenario 1} & 0.67$kg$ & 1$kgm^2$ & 0.25 & 5 & 1 & 1 \\ \hline
    \end{tabular}
    \label{table:cp1}
\end{table}
\begin{table}[t!]
    \centering
    \caption{Robot and Control Parameters}
    \begin{tabular}{| c | c | c | c | c | c | c |c|} 
        \hline
        \textbf{Robot Para.}  & $n_1$ & $n_2$  & $k$ &$f_{max}$ &$-$ &$-$ &$-$ \\ 
        \hline
        \textbf{Values}  & 1 & 1   & 400  & 4N &- &- &- \\ \hline

        \textbf{Control Para.} &  $f_0$ & $K_{Po}$ & $K_{Io}$ & $K_{Do}$   & $K_\omega$ & $K_{P,in}$ &  $K_{I,in}$  \\ 
        \hline
        \textbf{Values} &1.5N  & 3 & 0.5 & 0  &10 & 20 & 4 \\ \hline
    \end{tabular}
    \label{table:cp2}
\end{table}
\subsection{Simulation and Results}
\begin{figure}[t!]
\begin{subfigure}[t]{\linewidth}
\centerline{\includegraphics[ width=\linewidth,height=0.35\linewidth]{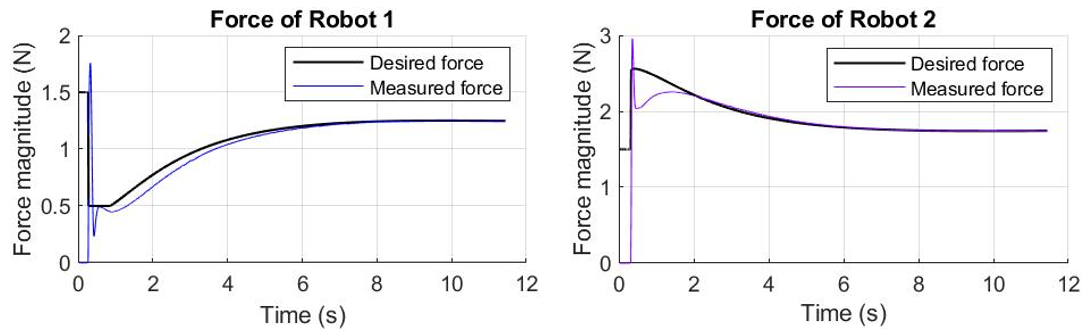}}
\caption{The convergence of estimated forces of two robots to desired forces.}
\end{subfigure}
\begin{subfigure}[t]{0.49\linewidth}
\centerline{\includegraphics[width= \linewidth,height=0.7\linewidth]{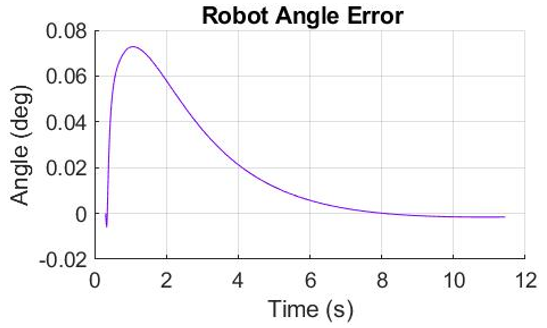}}
\caption{Identical convergence of contact angles $\theta_i$ (robot heading errors) for Robots 1 and 2 to zero.}
\end{subfigure}
\begin{subfigure}[t]{0.49\linewidth}
\centerline{\includegraphics[width= \linewidth,height=0.75\linewidth]{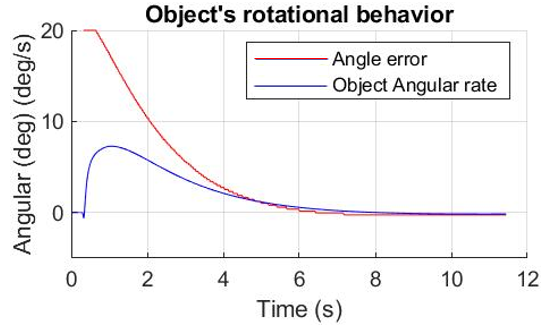}}
\caption{The convergence of motion direction error $\alpha_i$ and object's angular velocity $\omega$ to zero.}
\end{subfigure}
\caption{Simulation results demonstrating the convergence to zero.}
\label{fig_simu2}
\end{figure}

We conducted a Matlab simulation to model the behaviour of robots and objects based on the rigid object dynamics equations, differential drive robot models, and our proposed control DCFC. The simulation results are depicted in Fig. \ref{fig_simu2} and the stability and convergence of both the inner and outer loop control objectives were validated. Regarding the inner loop control, the measured force consistently converged towards the desired force signal, as shown in Fig. \ref{fig_simu2}a, represented by Eq. \ref{eq_in1}. In Fig. \ref{fig_simu2}b, the contact angles $\theta_i$ of both robots also converge to zero as in Eq. \ref{eq_in2}. As shown in Fig. \ref{fig_simu2}c, the object's motion direction is closely aligned with the desired direction, meaning the angle error converges to zero. Meanwhile, the object's rotational behavior remains stable as its angular velocity approaches zero. Note that the settling time of the entire system is approximately 6 seconds.
\vspace{-4mm}
\subsection{Real-world experiments and Results }
The real-world experiment setup details are outlined in our previous work \cite{ViART}. Each robot has a Zumo platform and a Raspberry Pi 4 mini-computer, enabling individual decision-making based on local contact force and angle data. These robots do not use inter-agent communication capabilities and solely receive desired directions in their robot frames thanks to the motion-tracking system. Notably, an alternative sensor like a digital compass could replace the motion tracking system's signal in practice. The coordinate of the moving object is only collected for system performance evaluation. We investigate two distinct scenarios for evaluating control performance: case study 1 with a constant desired direction and case study 2 with a changing desired direction.
\begin{figure}[t!]
\begin{subfigure}[t]{\linewidth}
\centerline{\includegraphics[width= 1.04\linewidth]{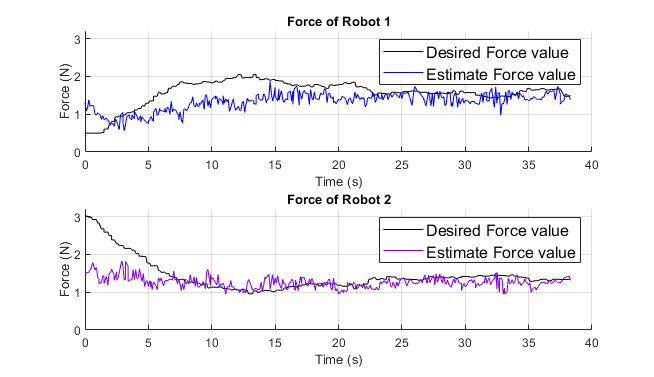}}
\caption{The convergence of estimated forces of two robots to desired forces.}
\end{subfigure}
\begin{subfigure}[t]{\linewidth}
\centerline{\includegraphics[width= 1.04\linewidth]{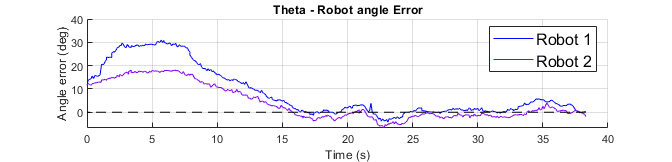}}
\caption{The convergence of contact angles $\theta_i$ (robot heading errors) for robots 1 and 2 to zero.}
\end{subfigure}
\begin{subfigure}[t]{\linewidth}
\centerline{\includegraphics[width= 1.04\linewidth]{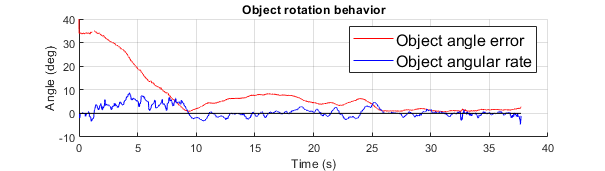}}
\caption{Object rotation captured by a motion capture and tracking system demonstrating the convergence of motion direction error $\alpha_i$ and object's angular velocity $\omega$.}
\end{subfigure}

\caption{Real-world experiment results for case 1 with constant desired moving direction.}
\label{fig_result1}
\end{figure}

\begin{figure}[t!]
\begin{subfigure}[t]{\linewidth}
\centerline{\includegraphics[width= 1.07\linewidth]{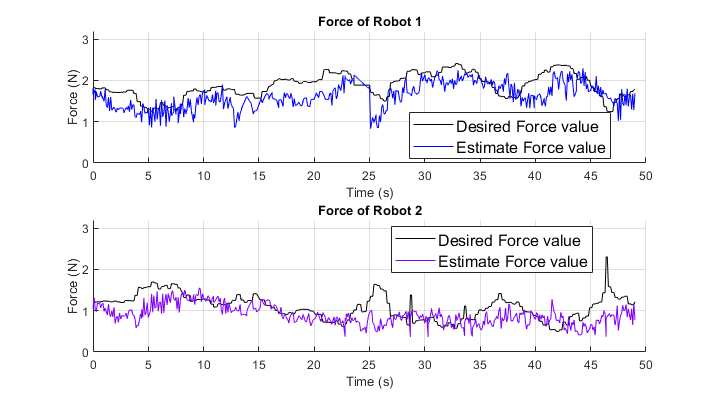}}
\caption{The convergence of estimated forces of two robots to desired forces.}
\end{subfigure}
\begin{subfigure}[t]{0.99\linewidth}
\centerline{\includegraphics[width= 1.02\linewidth]{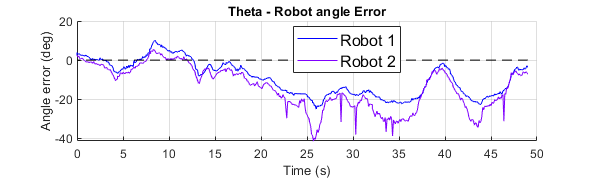}}
\caption{The convergence of the contact angles $\theta_i$ (robot heading errors) for robots 1 and 2 to zero.}
\end{subfigure}
\caption{Real-world experiment results for case 2 with changing desired moving direction.}
\label{fig_result2}
\end{figure}

In our real-world experiments\footnote{Real-world experiment results: https://youtu.be/tBqVyBXuhgU}, the stability and convergence of both the inner and outer loop control objectives were also validated in both scenarios. For the inner loop control, we observed that the force applied by both robots fluctuated about the desired force signal, as depicted in Fig. \ref{fig_result1}a for case study 1 and Fig. \ref{fig_result2}a for case study 2. After a 10-second interval, the mean absolute steady-state errors (MAE) in force for Robots 1 and 2 in case study 1 were 0.28N and 0.13N, respectively. In case study 2, the force tracking MAE values were 0.30N for both robots. Considering that the forces exerted by the robots were consistently non-negative, this ensured continuous pushing contact between each robot and the object during transportation tasks. Furthermore, robot heading errors $\theta_1$ and $\theta_2$ tended to be approximately close to zero for case study 1 (Fig. \ref{fig_result1}b) and case study 2 (Fig. \ref{fig_result2}b). For the outer loop control objectives, the object's direction of motion exhibited only minimal deviation from the desired value, as illustrated in Fig. \ref{fig_result1}c. Moreover, the object's rotational behaviour remained stable, with its angular velocity approximately zero.

\subsection{Discussion}
Our simulation and real-world experiments have validated the reasonable convergence time and stability of our proposed DCFC control system. Certain system parameters, particularly those related to friction, are subject to uncertainty because the surface conditions in real-world experiments do not remain as constant as they are in the model. The hardware results collectively confirm that our proposed DCFC control scheme successfully overcame these uncertainties, enabling robots to transport the object to the moving target.

Although the control policy can handle unknown object mass and friction parameters, $f_0$ and the number of robots are pre-defined to counteract the static friction between the object and the surface. The exact number of robots and their positions are not critical as long as there is sufficient space on the object's boundary for them. This setup can be exemplified by positioning two robots on the left and right edges of the object. In the near future, we aim to deploy more robots to transport objects with indeterminate masses and to handle asymmetric pushing positions, thereby assessing the robustness of the designed control system. Moreover, we plan to introduce a pre-pushing phase where robots determine suitable pushing positions using their local sensors for more effective object manipulation. The control system currently does not account for the specific shape of the object, provided that the shape allows the distribution of the robots on two sides of the center of mass along a straight edge, such as in triangles or certain polygons. In subsequent developments, we will enhance the algorithm to cater to a wider range of object shapes, especially those with non-polygonal geometries and non-uniform mass distributions.

\section{CONCLUSIONS}
In conclusion, our proposed distributed cascade force control system (DCFC) helps soft-tactile-based robots push an unknown object in a desired direction without communication. Moreover, robots can successfully exert desired force and maintain pushing contact while transporting objects using deformable tactile sensors. The two-loop cascade control system demonstrated coordination capacity among robot groups, enabling precise object transportation with proven convergence and stability theory. The insights gained from this research will advance the field of cooperative object transportation with soft-tactile-based robots, leading to more robust and versatile robotic systems for real-world applications.

\addtolength{\textheight}{-12cm}   


\bibliographystyle{IEEEtran}
\bibliography{Reference}

\end{document}